\documentclass[10pt,twocolumn,letterpaper]{article}

\usepackage{cvpr}              
\usepackage{algorithm}
\usepackage{algorithmic}
\usepackage[dvipsnames]{xcolor}

\definecolor{cvprblue}{rgb}{0.21,0.49,0.74}
\usepackage[pagebackref,breaklinks,colorlinks,citecolor=cvprblue]{hyperref}

\usepackage{float}
\usepackage{multirow}
\usepackage{graphicx}
\usepackage{colortbl}
\usepackage{utfsym}
\usepackage{subcaption}
\usepackage{tabularx}
\usepackage{caption}
\usepackage{wrapfig}
\usepackage{graphicx}
\usepackage{multirow}
\usepackage{colortbl}
\usepackage{booktabs}
\usepackage{amssymb}
\usepackage{bbding}
\usepackage{pifont}
\usepackage{wasysym}
\usepackage{fontawesome}
\usepackage[export]{adjustbox}

\usepackage{pifont}

\definecolor{sh_blue}{rgb}{0,0.60,0.93}
\definecolor{sh_gray2}{rgb}{1,0.89,0.75}
\definecolor{color3}{rgb}{0.95,0.95,0.95}
\definecolor{mygray}{gray}{.9}
\definecolor{bluegreen}{rgb}{0.44, 0.64, 0.77}
\definecolor{gray_venue}{rgb}{0.53,0.52,0.52}
\definecolor{color5}{rgb}{1,0.96,0.88}

\newlength{\Oldarrayrulewidth}

\usepackage[accsupp]{axessibility}  

\def\paperID{***} 
\def\confName{CVPR}
\def\confYear{2024}

\title{Efficient Dataset Distillation through Low-Rank Space Sampling}

\author{
Hangyang Kong$^{1,2}$ \quad Wenbon Zhou$^{1,3}$ \quad Xuxiang He$^{1,3}$ \quad Xiaotong Tu$^{1,2,3}$ \quad Xinghao Ding$^{1,2,3}$\thanks{Xinghao Ding (dxh@xmu.edu.cn) is the corresponding author.}\\ \vspace{-0.5mm}
  $^{1}$Key Laboratory of MultimediaTrusted Perception and EfficientComputing, \\Ministry of Education ofChina, Xiamen University\quad \\
  $^{2}$Institute of Artificial Intelligence, Xiamen University\quad \\ \vspace{-0.5mm}
  $^{3}$School of Informatics, Xiamen University\quad \\ \vspace{-0.5mm}
}

\begin{document}






\maketitle

\begin{abstract}
Huge amount of data is the key of the success of deep learning, however, redundant information impairs the generalization ability of the model and increases the burden of calculation. Dataset Distillation (DD) compresses the original dataset into a smaller but representative subset for high-quality data and efficient training strategies. Existing works for DD generate synthetic images by treating each image as an independent entity, thereby overlooking the common features among data. This paper proposes a dataset distillation method based on Matching Training Trajectories with Low-rank Space Sampling(MTT-LSS), which uses low-rank approximations to capture multiple low-dimensional manifold subspaces of the original data. The synthetic data is represented by basis vectors and shared dimension mappers from these subspaces, reducing the cost of generating individual data points while effectively minimizing information redundancy. The proposed method is tested on CIFAR-10, CIFAR-100, and SVHN datasets, and outperforms the baseline methods by an average of 9.9\%.

\end{abstract}
\vspace{-0.6cm}

\section{Introduction}
\vspace{-0.1cm}
\label{sec:intro}
With the development of deep learning\cite{lecun2015deep}, large-scale datasets have become essential for training accurate and robust models. However, the growing size of datasets significantly increases information redundancy and training costs. Dataset distillation (DD) addresses this challenge by synthesizing high-information-density small datasets to enhance training efficiency.
Dataset distillation addresses this challenge by synthesizing small, high-information-density datasets, reducing the burden of data storage and transmission while improving training efficiency. Dataset distillation has numerous valuable applications, including continuous learning\cite{cubuk2019autoaugment,rosasco2021distilled}, neural architecture search\cite{zoph2016neural,wang2021rethinking}, and privacy-preserving tasks\cite{zhao2021dataset}.

Dataset distillation was first proposed by Wang\cite{wang2018dataset}, aiming to extract knowledge from large datasets to create smaller synthetic datasets that achieve comparable performance to models trained on the original datasets. Over time, methods such as gradient matching algorithms\cite{zhao2020dataset,zhang2023accelerating,liu2023dream}, kernel-induced point\cite{nguyen2020dataset}, distribution matching strategies\cite{wang2022cafe}, and trajectory matching method\cite{cazenavette2022dataset,guo2023towards} have been proposed as the primary methods of dataset distillation.
The trajectory matching method, currently the mainstream approach, guides the generation of synthetic data by comparing the similarity of loss trajectories during model training, thereby better preserving the consistency of the training process.
Despite the promising results achieved by the aforementioned methods, they neglect the similar characteristic information across different data. This oversight can result in information redundancy among the synthetic data. When the Images Per Class(IPC) is small, this drawback significantly affects the quality of the synthetic data.

\begin{figure*}[t]
\centering
\includegraphics[width=0.88\textwidth]{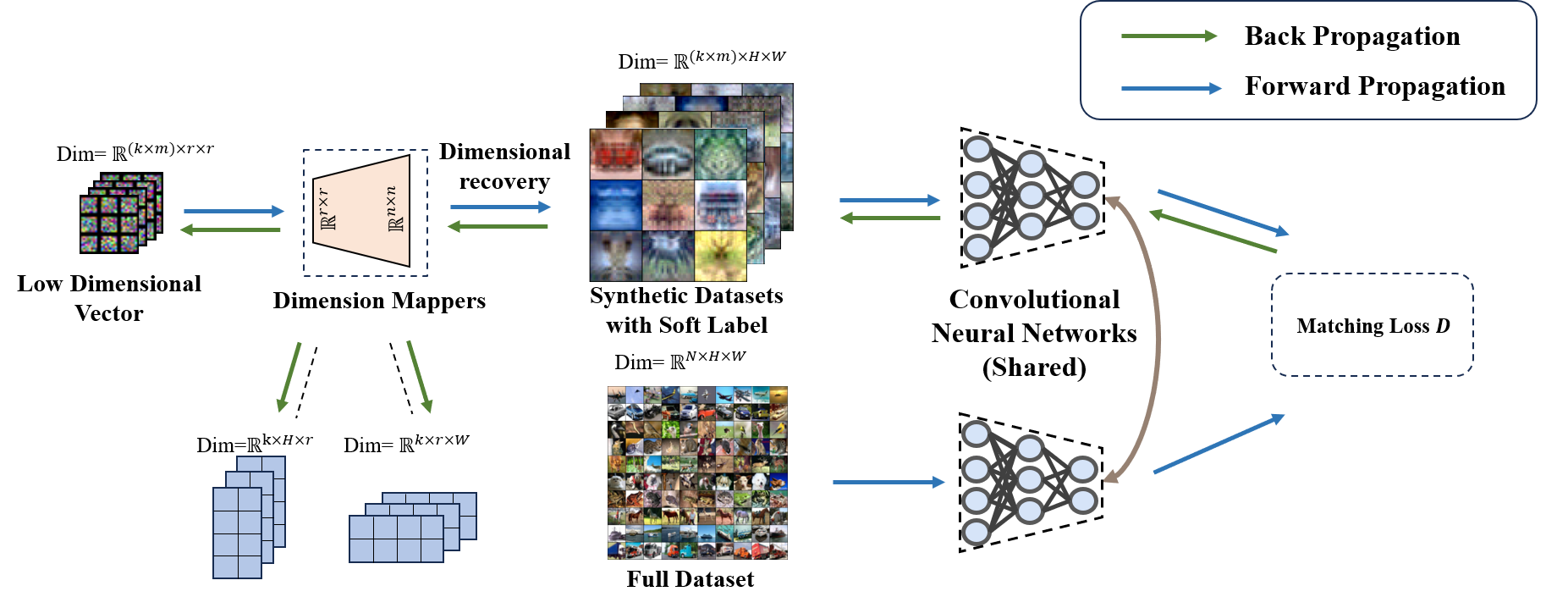} 
\caption{Illustration of the dataset distillation framework based on low-rank space sampling. The synthetic dataset is composed of basis vectors and dimension mappers. The dimension mappers are shared components for multiple basis vectors, preserving some common features of the synthetic dataset and effectively reducing the information redundancy problem.}
\label{fig2}
\end{figure*}

In this paper, we propose a dataset distillation method based on Matching Training Trajectories with Low-rank Space Sampling(MTT-LSS). This method introduces a low-rank approximation technique to capture multiple low-rank manifold subspaces of the original dataset. Each low-dimensional manifold subspace is represented by a set of dimension mappers which map the original data dimensions into a low-rank space. Consequently, the synthetic data can be represented by the ``basis vectors'' of the low-dimensional subspaces. The dimension mappers, serving as shared matrices, preserve the similar characteristics of the synthetic dataset. Together with the basis vectors, they form a complete synthetic dataset, effectively addressing the issue of information redundancy present in previous methods. Additionally, to avoid the potential negative impact of preset hard labels on the optimization process of the dimension mappers, we propose initializing the labels with soft labels and treating them as ``meta-parameters'' that are optimized during the training process. Using soft labels provides a finer-grained non-target probability distribution, making it easier to find the optimal solution for the model.The optimization process of synthetic images can be seen in Fig.~\ref{fig2}.

In summary, our contributions are as follows:
\begin{itemize}
\item We propose a dataset distillation method based on low-rank space sampling with a shared matrix, aiming to reduce information redundancy among synthetic dataset and optimize the synthetic dataset.
\item We introduce soft labels to replace the original hard labels, providing a finer-grained non-target probability distribution.
\end{itemize}
\vspace{-0.2cm}

\section{Related Works}
\subsection{Dataset Distillation}
Dataset distillation can be interpreted as a meta-learning problem with inner and outer loop optimization\cite{hospedales2020metalearningneuralnetworkssurvey}, where the synthetic dataset \( S \) serves as the meta-parameter to be learned by the algorithm. In the inner loop, model parameters \( \theta \) are optimized based on the synthetic dataset \( S \), while in the outer loop, \( S \) is optimized by minimizing the meta-training loss, which incorporates information from the target dataset. The gradient of the meta-training loss is referred to as the meta-gradient.

Wang et al.\cite{wang2018dataset} first proposed the task of dataset distillation, using a one-step optimization approach to compute the meta-training loss. Sucholutsky and Schonlau\cite{Sucholutsky_2021} extended Wang et al.'s work by learning soft labels in the synthetic dataset for better information compression, treating both images and labels in \( S \) as meta-parameters.
Zhao et al. \cite{zhao2020dataset}proposed matching the network parameter changes trained on \( S \) and the target dataset \( T \) for dataset distillation, using class gradients as surrogate targets to bridge the gap between the two, thus extracting information from the original dataset.Lee et al.\cite{lee2022datasetcondensationcontrastivesignals} noted that class gradient matching overly focuses on common class features, neglecting discriminative features between classes. They proposed that the sum of inter-class loss gradients better captures the constraints needed for network training. Zhao and Bilen \cite{zhao2021dataset}highlighted that synthetic datasets do not benefit from data augmentation during training, and introduced DSA (Data Siamese Augmentation) to homogeneously augment both distilled and target data during the distillation process.

Unlike gradient matching methods, Cazenavette et al. \cite{cazenavette2022dataset}directly matched the long-range training trajectories between the original and synthetic datasets. They collected pre-trained network change trajectories on the original dataset as expert trajectories, randomly initialized the network \( \theta_S \) used to train \( S \) from these expert trajectories, and matched the training trajectories of \( S \) with the expert trajectories. Du et al. \cite{du2023minimizing}pointed out that due to differences in the real trajectories between \( S \) and \( T \), initializing \( \theta_S \) with expert trajectories can cause trajectory accumulation errors. They mitigated this error by adding random noise during the initialization of \( \theta_S \), thus improving robustness. Cui et al. \cite{cui2022dcbenchdatasetcondensationbenchmark}proposed benchmark evaluation methods for dataset distillation to better assess the quality of synthetic data. By decomposing the meta-gradient into two processes, Cui et al. \cite{cui2023scaling}significantly reduced the memory required for trajectory matching and successfully extended the trajectory matching method to the large-scale ImageNet-1k dataset.

\subsection{Low-rank Method}
Low-rank approximation methods \cite{liu2012robust}often involve convex optimization problems that are easy to solve globally and their stability and convergence results usually do not depend heavily on initialization. Therefore, this approach is common in machine learning applications such as computer vision\cite{Pope_Zhu_Abdelkader_Goldblum_Goldstein_2021}, natural language processing\cite{Aghajanyan_Gupta_Zettlemoyer_2021,J._Shen_Wallis_Allen-Zhu_Li_Wang_Chen_2021}, and recommender systems\cite{Koren_Bell_Volinsky_2009,yang2023efficient,lin2024aglldiff,lin2024unsupervised,lin2023domain,lin2024fusion2void,wang2023learning,wang2025learning,wang2024cross,wang2025towards,zhang2024frequency,huang2023dp,chen2023cplformer,chen2024teaching} also proposed that image data can be approximated using low-rank matrices, introducing low-rank decomposition methods into the dataset distillation task can effectively reduce the cost of obtaining a single synthetic image, providing an orthogonal method to existing dataset distillation approaches. However, this method optimizes synthetic data samples independently, without considering the similarity between data. Our proposed method addresses this drawback.

\section{Preliminaries}
Using dataset distillation to obtain synthetic data, an important objective is to reduce the acquisition cost of individual data samples and achieve more compact feature representations. There is existing evidence demonstrating that data augmentation techniques can significantly improve dataset distillation tasks from a gradient-based perspective\cite{zhou2024dataset}. Given the original large dataset \( T \), the goal is to minimize the distance function \( D(\cdot, \cdot) \) between the synthetic dataset \( S \) and \( T \) through objective matching. The proxy target acquisition algorithm for matching the data information between \( S \) and \( T \) in the dataset distillation task can be defined as \( Alg(S, l, \theta) \). Based on the quantity augmentation method, the number of synthetic data \( |S| \) is often increased through a deterministic and differentiable intermediate process \( f(\cdot) \). In the optimization process of synthetic data, \( f(S) \) replaces the original \( S \).Given the learning rate \( \eta \), the optimization update process for synthetic data is as follows:

\begin{equation}
S \leftarrow S - \eta \nabla_S D(Alg(f(S), l, \theta), T) 
\end{equation}

Based on the chain rule, the gradient of the matching distance loss function \( D(\cdot) \) with respect to the synthetic data \( S \) can be obtained through backpropagation:

\begin{equation}
\frac{\partial D(\cdot)}{\partial S} = \frac{\partial D(\cdot)}{\partial S_{enl}} \frac{\partial S_{enl}}{\partial f(\cdot)} \frac{\partial f(\cdot)}{\partial S} 
\end{equation}
where \( S_{enl} \) represents the expanded synthetic data obtained based on the quantity augmentation function \( f(\cdot) \). As \( |S| \) increases, the loss value in the target matching task decreases, and the test accuracy of the synthetic data improves\cite{kim2022dataset}. Therefore, for the information metric function \( I(\cdot) \), it can be considered that:

\begin{equation}
I\left( \frac{\partial D(\cdot)}{\partial S_{enl}} \right) > I\left( \frac{\partial D(\cdot)}{\partial S} \right), \quad |S_{enl}| > |S| 
\end{equation}

Based on equation (3), the optimization process of synthetic data can be written as:

\begin{equation}
S \leftarrow S - \eta \nabla_{S_{enl}} D(S_{enl}, T) \frac{\partial S_{enl}}{\partial f(\cdot)} \frac{\partial f(\cdot)}{\partial S}
\end{equation}

That is, based on the quantity augmentation function \( f(\cdot) \), more informative gradient information can be used to update the synthetic data. At high compression rates, a limited amount of image data can't encapsulate the diverse information present in the original dataset. Data augmentation methods address this issue by reducing redundancy within the synthetic data without increasing storage costs. 

\begin{figure}[t]
\centering
\includegraphics[width=1\columnwidth]{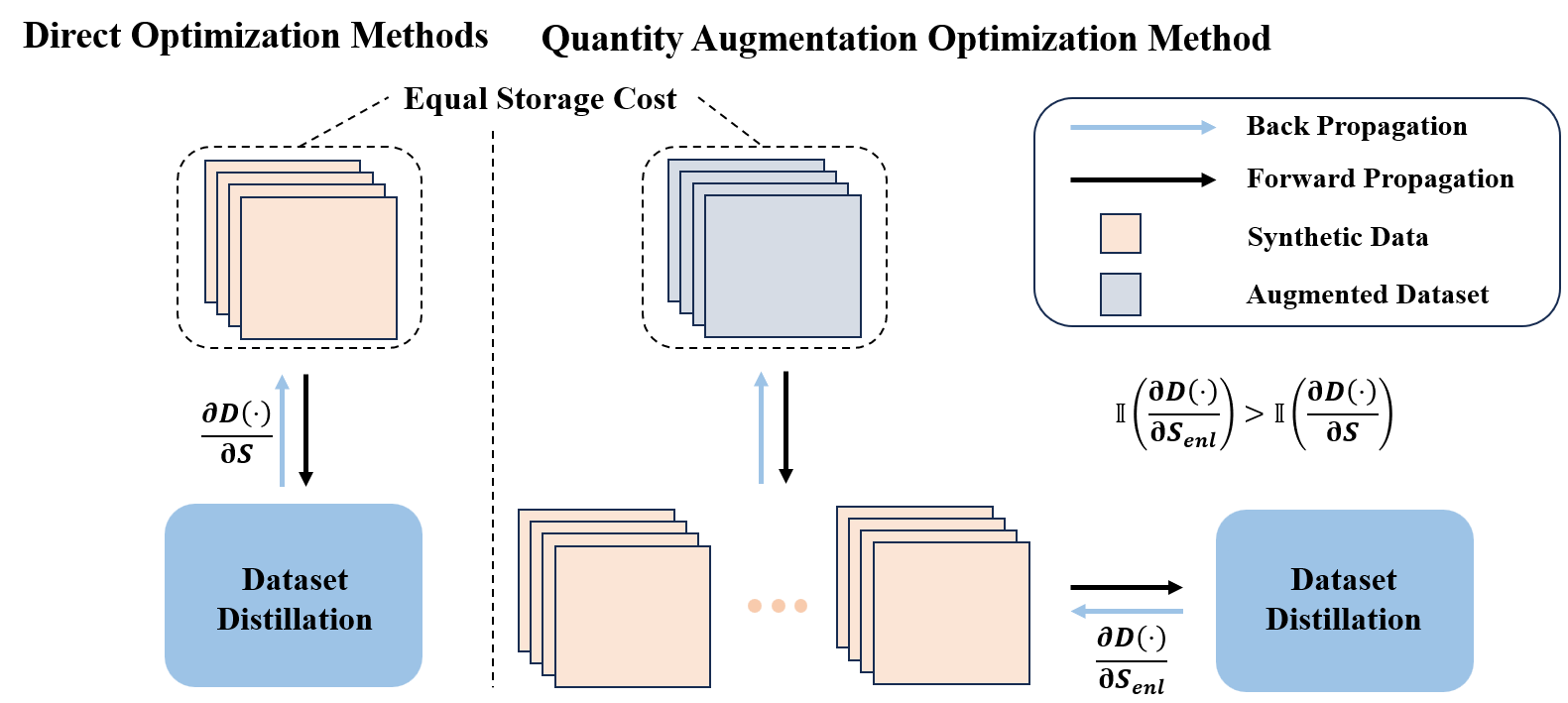} 
\caption{Comparison of Conventional Data Distillation Paradigm and Data Augmentation Distillation Paradigm. By employing data augmentation methods, a more compact feature representation can be achieved, enabling the generation of a larger number of synthetic images at the same storage cost.}
\label{fig1}
\end{figure}

\section{Methods}
\subsection{Low-Rank Space Sampling Method}

Introducing low-rank decomposition methods into the dataset distillation task can effectively reduce the cost of obtaining a single synthetic image. By singular value decomposition, a single image can be approximated by a low-rank matrix. 
For an image data \( x_i \in \mathbb{R}^{H \times W} \), it can be decomposed as:

\begin{equation}
    x_i = U \Sigma V^T 
\end{equation}
where \( U \) is an \( H \times H \) unitary orthogonal matrix, with its column vectors called left singular vectors; \( \Sigma \) is an \( H \times W \) diagonal matrix, with its diagonal elements being singular values, usually arranged in descending order representing the ``strength'' in the direction of the singular vectors; \( V^T \) is the transpose of a \( W \times W \) unitary orthogonal matrix, with its row vectors called right singular vectors. In practical applications, the singular values of an image usually decay rapidly, meaning only the first \( r \) singular values are relatively large, while the rest are relatively small. This indicates that the main information of the image is concentrated in the first \( r \) singular values and their corresponding singular vectors. Therefore, a low-rank approximation of the original image matrix can be obtained by retaining only the primary information part:
\begin{equation}
    x_i \approx U_{(r)} \Sigma_{(r)} V_{(r)}^T 
\end{equation}
where \( U_{(r)} \) is the first \( r \) columns of \( U \), forming an \( H \times r \) matrix; \( \Sigma_{(r)} \) is a diagonal matrix consisting of the first \( r \) singular values; \( V_{(r)}^T \) is the first \( r \) rows of \( V^T \), forming an \( r \times W \) matrix. In this paper, multiple low-rank approximated data samples can be normalized into a set of dimension mappers, thus obtaining the subspace of the original data in the low-dimensional manifold. Then, basis vector sampling is performed in the low-dimensional space at a low cost, and the basis vectors can complete the data dimension recovery based on the dimension mappers. Specifically, the proposed method uses multiple sets of dimension mappers \(\{U_i, V_i^T\}_{i=0}^k\) and basis vectors \(\{\Sigma_i\}_{i=0}^{k \times m}\) in the low-dimensional manifold space (where \( U_i \in \mathbb{R}^{H \times r} \); \( V_i^T \in \mathbb{R}^{r \times W} \); \( \Sigma_i \in \mathbb{R}^{r \times r} \); \( k \) is the number of dimension mapper sets; \( m \) is the number of basis vectors sampled in each low-dimensional manifold space). Each set of ``dimension mappers'' can map multiple ``low-dimensional basis vectors'' to the original data dimension space, forming a set of representative synthetic data:

\begin{equation}
    x_i = U_i \Sigma_i V_i^T 
\end{equation}

The synthetic data \( S = \{x_i\}_{i=0}^{k \times m} \), based on Equation (1), can have its optimization objective defined as:

\begin{equation}
    S^* = \arg \min_S D(Alg(U, \Sigma, V, l, \theta), T) 
\end{equation}

\subsection{Soft Labels and Progressive Optimization Strategy}

Due to the unclear nature of the subspace information mapped by the dimension mappers before optimization, it is challenging to determine the optimal initialization of label data $\{y_i\}_{i=0}^{k \times m}$ for the subspace optimization process. Hence, assigning fixed hard labels to all synthetic data at the outset might adversely affect the optimization of the dimension mappers. For example, if the synthetic data labels under a particular "dimension mapper" are set to the same category, the subspace formed by that "dimension mapper" can be considered the location of that specific category. This pre-determined data label might not be the optimal solution for the dataset distillation task. To find the optimal labels during the optimization process, our method randomly initializes the label information $\{y_i\}_{i=0}^{k \times m}$ of the synthetic dataset and treats these labels as "trainable meta-parameters," optimizing them using gradient information from the dataset distillation task to find the optimal solution.

The use of trainable soft labels has two advantages:

\begin{itemize}
    \item  It avoids the interference of hard label initialization on the optimization process of the dimension mappers. Soft labels can be updated and optimized based on gradient information obtained during the dataset distillation process, thereby finding a set of optimal data labels that best fit the dimension mappers.
    \item  Soft labels provide additional inter-class relationship information for model training. Compared to hard labels, soft labels can more accurately express the relative relationships and probability distributions between classes. This information is crucial for the model to understand the feature distribution of different class data, significantly enhancing the model's generalization ability.
\end{itemize}

In the early stages of deep learning model training, parameter changes primarily capture basic features of the dataset, often leading to rapid improvements in model accuracy. However, in the later stages, model accuracy tends to oscillate, which can be attributed to fitting more difficult samples in the dataset. Based on this theory, we can divide expert trajectories into two parts according to the ease of feature fitting: simple trajectories $\theta_{easy}$ obtained in the early stages of network training, and hard trajectories $\theta_{hard}$ obtained in the later stages of network training.

In the trajectory matching method MTT, the authors use a hyperparameter to preset the maximum initial epochs of expert trajectories, which remains constant throughout the training process. This approach does not consider the optimization state of the current synthetic dataset, applying the same constraint information to the synthetic dataset at different optimization stages rather than dynamically adjusting based on its optimization state. This can lead to:

\begin{itemize}
    \item During the early training epochs of the dataset distillation task, the synthetic data $S$ captures very little feature information and may retain a lot of noise from the initialization phase. If hard trajectories $\theta_{hard}$ are used for trajectory matching optimization at this stage, the difficulty of feature learning far exceeds the current optimization state of the synthetic data, leading to a decrease in the accuracy of $S$, and even causing gradient explosion.
    \item Conversely, during the later training epochs of the dataset distillation task, the synthetic data has already learned the basic features of the original dataset well. If simple trajectories $\theta_{easy}$ are still used for trajectory matching optimization, the synthetic data will not learn valuable information and may overfit to the simple features.
\end{itemize}

\begin{figure*}[t]
\centering
\includegraphics[width=0.8\textwidth]{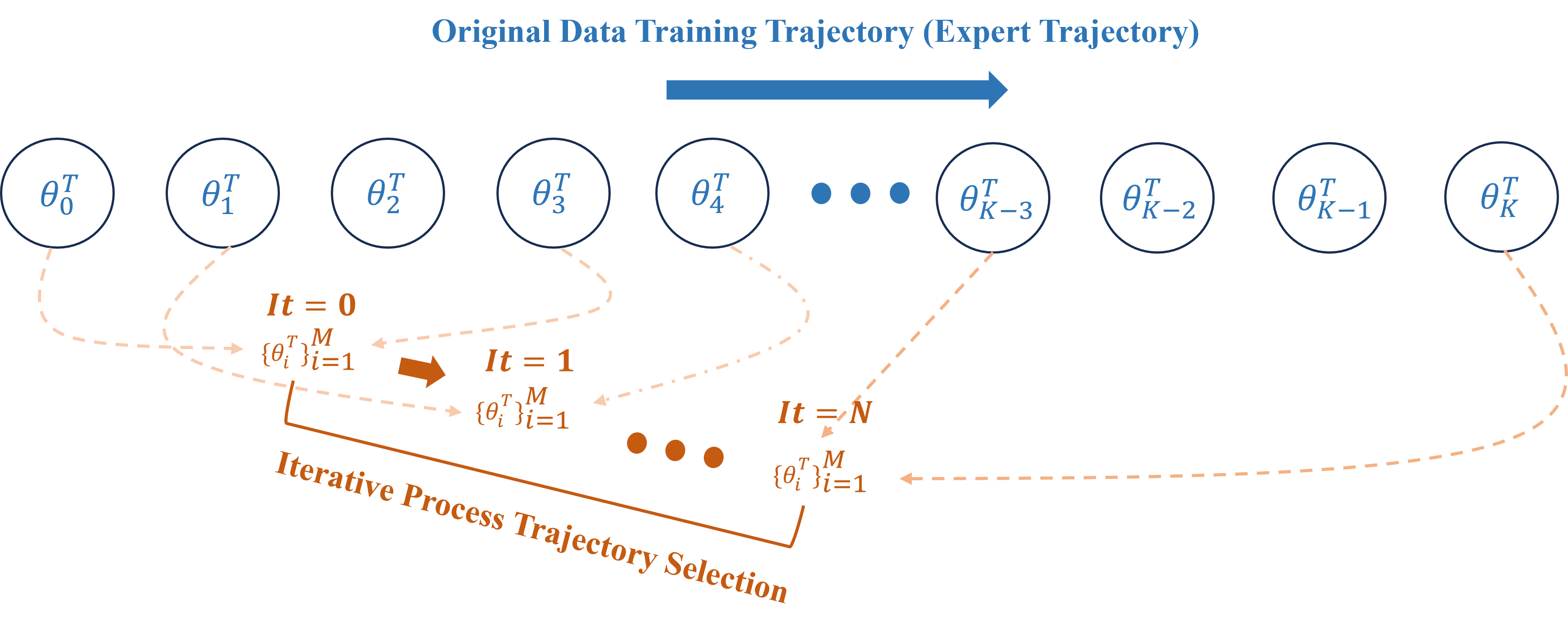} 
\caption{Using a step-by-step optimization strategy enables synthetic data to learn simple features at the beginning of training, and gradually increases the difficulty of matching trajectories as the feature representation capability of synthetic data increases.}
\label{fig3}
\end{figure*}

To address these issues, we adopt a progressive optimization strategy to obtain the expert trajectories to be matched, as shown in Fig 3. Specifically, considering the learning ability of the synthetic dataset, we gradually increase the initial epochs $i$ of the expert trajectories $\theta_i^T$ during the training iterations:

\begin{equation}
MaxStart_{it} = \frac{it}{Epochs} \cdot MaxStart + \delta
\end{equation}

\begin{equation}
Start_{it} = rand(MaxStart_{it} - w, MaxStart_{it})
\end{equation}
where $it$ denotes the current iteration, $\delta$ is a given minimum value for the expert's initial epochs, and $w$ is the floating value when obtaining the initial epochs. This strategy aims for the synthetic data to first learn simple features during the early training phase, and gradually increase the difficulty of matching trajectories as the synthetic data's feature representation capability improves.
The overall training process using trajectory matching as the baseline method is shown in Algorithm 1.

\begin{algorithm}[h]
\caption{Trajectory Matching Algorithm Based on Low-Rank Space Sampling (LS-MTT)}
\label{alg:ls-mtt}
\textbf{Input}: large-scale dataset $T$, expert trajectory length $M$, student trajectory length $N$, differentiable augmentation function $A$, low-rank value $r$, maximum start epoch of expert trajectory $maxStartEpoch$\\
\textbf{Output}: Synthetic dataset $S = (U, \Sigma, V)$, soft label set $L$
\begin{algorithmic}[1]
\STATE Initialize dimension mapping functions, base vectors, and soft labels: $U, V, \Sigma, L$
\FOR{$it = 1$ to $epoch_{max}$}
    \STATE Choose the initial epoch $ startEpoch_{it} $ based on Eq. 10
    \STATE Sample expert trajectories $\{ \theta_i^T \}_{i=0}^N$ from the buffer based on $startEpoch_{it}$
    \STATE Initialize the student network: $\theta_0^S \gets \theta_0^T$
    \FOR{$n = 0$ to $N-1$}
        \STATE  Smaple a mini-batch of samples from the synthetic data and their corresponding soft labels
            \STATE \quad $ batch_n \sim (U, \Sigma, V), \, l_n \sim L $
        \STATE Update the student network parameters based on classification loss
            \STATE \quad $\theta_n^S = \theta_{n-1}^S - \alpha \nabla l(A(batch_n), l_n; \theta_{n-1}^S)$
    \ENDFOR
    \STATE Compute the matching loss between student parameters and expert parameters:
        \STATE \quad $L = \| \theta_N^S - \theta_M^T \|_2^2/\;\| \theta_0^T - \theta_M^T \|_2^2$
    \STATE Update $U, \Sigma, V, L$ based on the learning rate
\ENDFOR
\end{algorithmic}
\end{algorithm}
\section{Experiments}
\subsection{Datasets and Experimental Setup}
\begin{table*}[t]
    \caption{
    Comparison of performance with state-of-the-art methods. 
     For other comparison methods, we extract a given number of images per class from the training set. The neural network is trained on the synthetic set and evaluated on the test set. LD and DD use AlexNet, while the rest use ConvNet for training and testing. Here, we use the MTT baseline as the matching method. IPC: number of images per class; Ratio (\%): proportion of extracted images to the entire training set.
    }
\label{table1}
\centering
\resizebox{\linewidth}{!}
{
\begin{tabular}{ccc|cccccccccccccc}
\hline
\multirow{2}{*}{} & \multirow{2}{*}{IPC} & \multirow{2}{*}{Ratio \%} & \multicolumn{14}{c}{Method}\\
& & & Random & Herding & DD & LD & DC & DSA & DM & CAFE & IDC & FRePo & MTT & HaBa & LoMTT & MTT-LSS(Ours)\\
\hline
\multirow{3}{*}{SVHN} & 1 & 0.2 & 14.6 & 20.9 & 23.7 & 25.1 & 31.2 & 27.5 & 38.7 & 42.9 & 68.1 & - & 58.5 & 69.8 & 65.7 & \textbf{78.4(+19.9)}\\
& 10 & 0.14 & 35.1 & 50.5 & 63.2 & 65.4 & 76.1 & 79.2 & 77.2 & 77.9 & \textbf{87.3} & - & 75.2 & 83.2 & 79.6 & 83.4(+8.2)\\
& 50 & 0.7 & 70.9 & 72.6 & 75.6 & 77.3 & 82.3 & 84.4 & 80.2 & 82.3 & 90.2 & - & 85.7 & 88.3 & 87.8 & \textbf{90.6(+4.9)}\\
\multirow{3}{*}{CIFAR-10} & 1 & 0.02 & 14.4 & 21.5 & 24.1 & 25.7 & 28.3 & 28.8 & 26.0 & 31.6 & 50.0 & 46.8 & 46.3 & 48.3 & 51.0 & \textbf{65.3(+19.0)}\\
& 10 & 0.2 & 26.0 & 31.6 & 36.8 & 38.3 & 44.9 & 52.1 & 48.9 & 50.9 & 67.5 & 65.5 & 65.3 & 69.9 & 67.8 & \textbf{71.3(+6.0)}\\
& 50 & 1 & 43.4 & 40.4 & 41.3 & 42.5 & 53.9 & 60.6 & 63.0 & 62.3 & 74.5 & 71.1 & 71.6 & 74.0 & 73.2 & \textbf{74.8(+3.2)}\\
\multirow{3}{*}{CIFAR-100} & 1 & 0.2 & 4.2 & 8.4 & 10.2 & 11.5 & 12.8 & 13.9 & 11.4 & 14.0 & 30.7 & 28.7 & 24.3 & 33.4 & 29.4 & \textbf{33.9(+9.8)}\\
& 10 & 2 & 14.6 & 17.3 & 21.0 & 22.7 & 25.2 & 32.3 & 29.7 & 31.5 & 44.8 & 42.5 & 40.0 & 40.2 & 41.3 & \textbf{46.3(+6.3)}\\
& 50 & 10 & 30.0 & 33.7 & - & - & 37.8 & 42.8 & 43.6 & 42.9 & - & 44.3 & 46.1 & 47.0 & 47.1 & \textbf{50.0(+3.9)}\\
\hline
\end{tabular}
}
\end{table*}
The experiments are conducted on three commonly used public benchmark datasets, including SVHN\cite{netzer2011reading}, CIFAR-10, and CIFAR-100\cite{krizhevsky2009learning}. The SVHN dataset comprises over 600,000 color digit images intended for digit recognition tasks against natural image backgrounds, with an image resolution of 32×32. The CIFAR-10 dataset consists of 60,000 32×32 color images across 10 categories, while the CIFAR-100 dataset contains 60,000 32×32 color images divided into 100 categories. These datasets are widely utilized in image recognition and image classification tasks.

To validate the superiority of the proposed method, this section compares the method with currently representative core-set acquisition methods and dataset distillation methods, including Random, Herding\cite{chen2012super}, DD\cite{wang2018dataset}, LD\cite{bohdal2020flexible}, DC\cite{zhao2020dataset}, DSA\cite{zhao2021dataset}, DM\cite{zhao2023dataset}, CAFE\cite{wang2022cafe}, IDC\cite{kim2022dataset}, MTT\cite{cazenavette2022dataset}, LoMTT\cite{yang2023efficient}, FRePo\cite{zhou2022dataset}, and HaBa\cite{liu2022dataset}. 

Our method employs low-rank spatial sampling for synthetic data, based on the MTT approach. In the comparative experiments, it ensures that the storage costs for the dimension mappers, the low-dimensional basis vectors, and the soft labels do not exceed the storage costs of the synthetic data generated by other methods.

\begin{figure}[t]
\centering
\includegraphics[width=1\columnwidth]{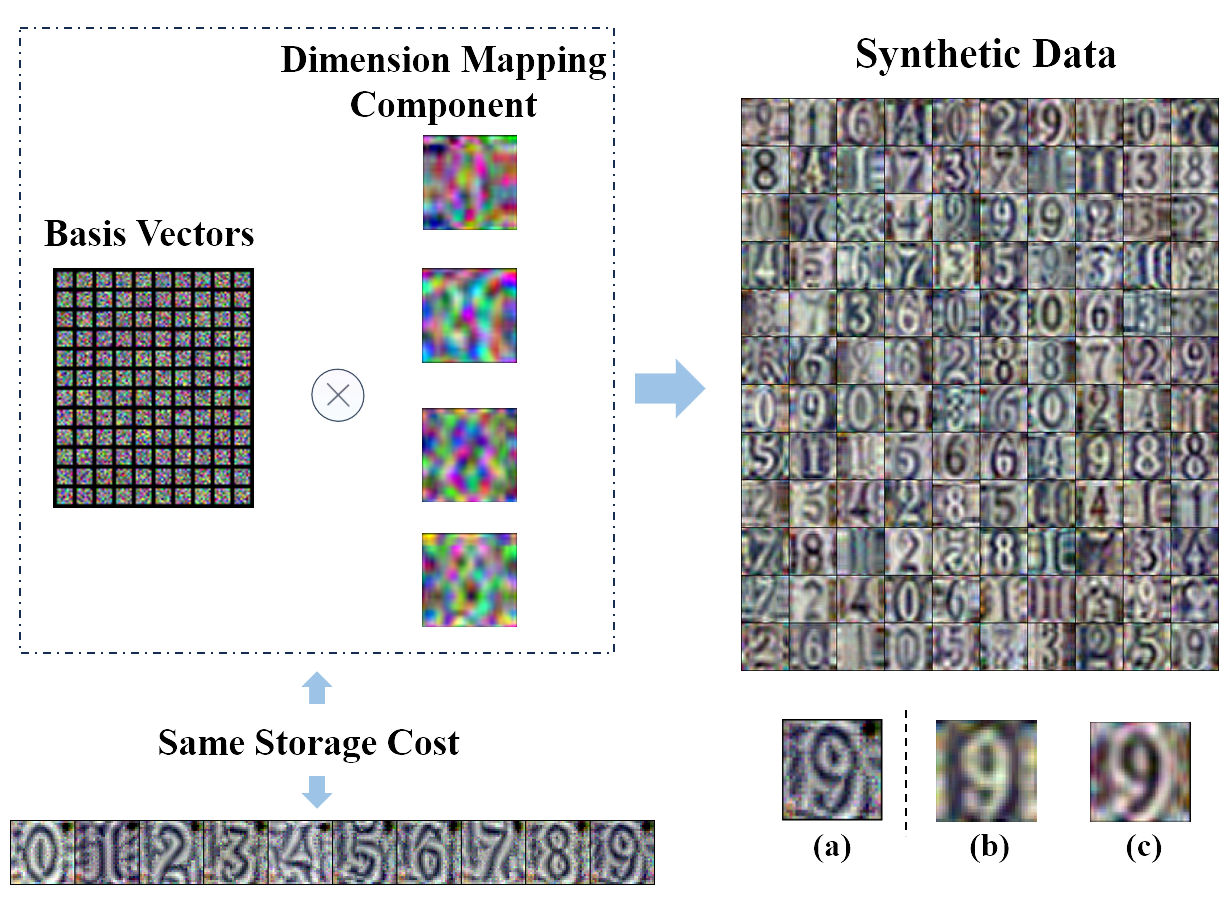} 
\caption{Dimension Mapping Component, Basis Vectors, and Visualized Results of the Synthesized Images (the lower part shows the results obtained by the baseline method). The low-rank sampling method based on the shared matrix can obtain more synthetic data at the same storage cost.}
\label{fig4}
\end{figure}

\begin{table}[htbp]
\centering
\begin{tabular}{cccc}
\hline
\(r \) & \(M \) & basis vectors & total size\\
\hline
4  & 15 & 15 × 22 & 30660\\
8  & 4  & 4 × 30  & 30384\\
12 & 4  & 4 × 12  & 30432\\
16 & 3  & 3 × 9   & 30222\\
\hline
\end{tabular}
\caption{Hyperparameter settings. \(M\): Number of low-dimensional components.}
\label{table2}
\end{table}

\begin{figure}[htbp]
\centering
\includegraphics[width=0.9\linewidth]{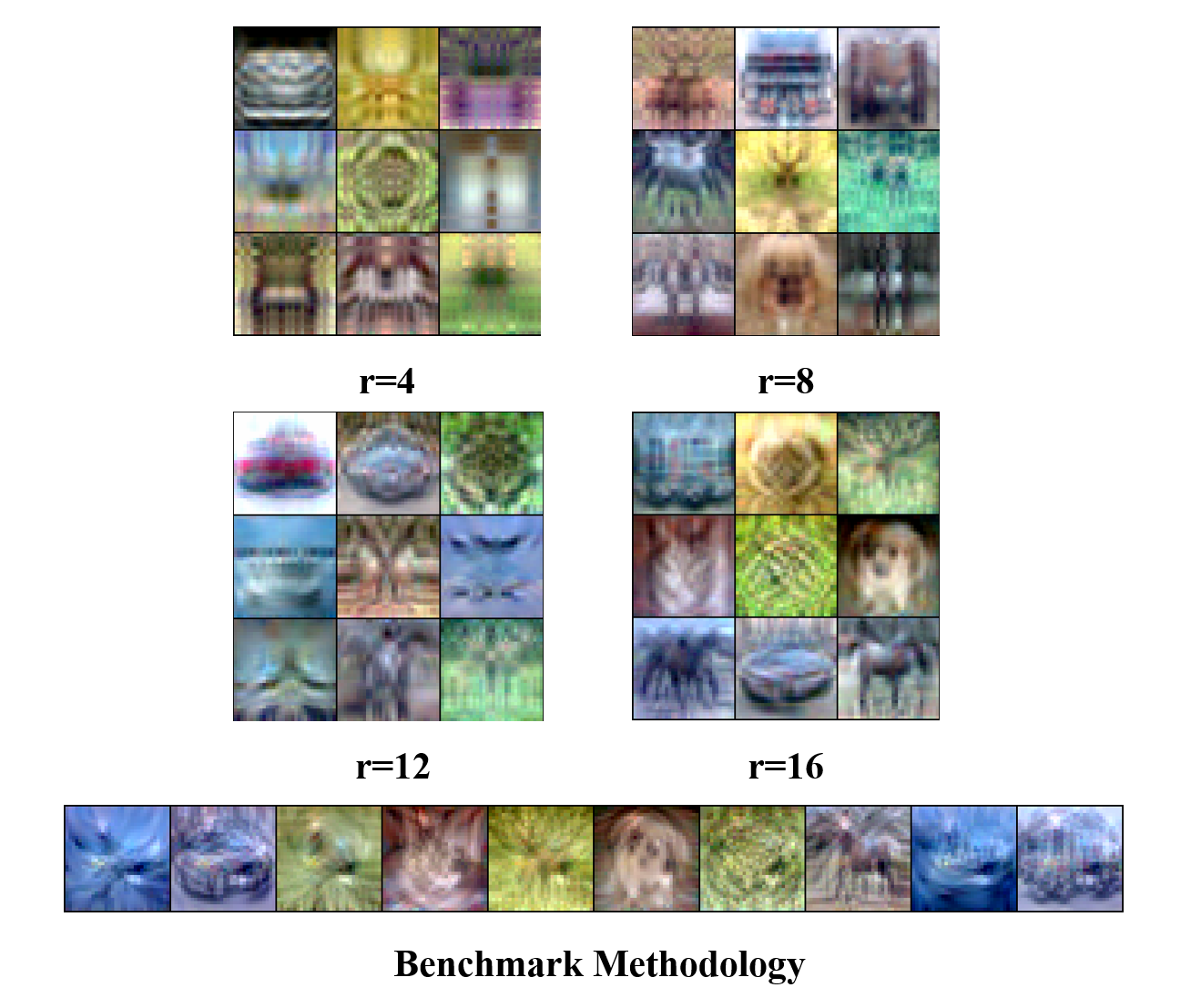} 
\caption{Comparison of synthesized images with different low-rank values \( r \). Higher \( r \) improves details but increases storage cost.}
\label{fig6}
\end{figure}

\begin{table}[htbp]
\centering
\small  
\setlength{\tabcolsep}{3pt}  
\begin{tabular}{ccc|cc}
\hline
Soft & Prog. & Low-rank & 0.02\% & 0.2\% \\
Labels & Opt. & Sampling & Ratio  & Ratio \\
\hline
 &  &  & 46.3 & 65.3\\
 & ✓ &  & 46.3 & 65.5\\
✓ &  &  & 48.1 & 66.2\\
✓ & ✓ &  & 48.1 & 66.7\\
 &  & ✓ & 63.2 & 69.8\\
 & ✓ & ✓ & 63.4 & 70.1\\
✓ &  & ✓ & 65.0 & 70.8\\
✓ & ✓ & ✓ & \textbf{65.3} & \textbf{71.3}\\
\hline
\end{tabular}
\caption{Ablation study on soft labels, progressive optimization, and low-rank sampling.}
\label{table3}
\end{table}

\subsection{Main Results}

In this section, the meta-test accuracy of our method is compared with all the comparative methods under different datasets and compression rates. The results are presented in Table 1, where IPC stands for the number of images per class for non-data augmentation methods. The results from the tables indicate that our method significantly improves the accuracy of the benchmark MTT method under various experimental settings. Furthermore, compared to all the dataset distillation methods, the proposed method achieves either the best or the second-best performance, which confirms the strong effectiveness of our method.

Upon analyzing the results in the tables, it is evident that our method demonstrates significant performance improvements under low compression rates. Specifically, when the IPC is set to 1, the method achieves a remarkable 19.9\% increase in meta-test accuracy on the SVHN dataset and a 19\% increase on the CIFAR10 dataset. This substantial improvement can be attributed to the fact that under low compression rates, the limited number of synthetic data severely restricts their ability to capture the diverse characteristics of the original data. However, our method addresses this limitation by employing a low-rank spatial sampling approach, which increases the number of synthetic data at the same storage cost. This effectively enhances the diversity of feature representation in the synthetic dataset, leading to superior meta-test performance.


As illustrated in Fig 4, it is observed that the dimension mappers stores some contour information of the data. The basis vectors appear to be noisy data but can be mapped to the SVHN data space based on the dimension mappers. Comparing the images obtained by our method with those from the benchmark method, it is evident that the background of the benchmark data contains noisy signals, whereas the images obtained by our method effectively filter out the noise. Additionally, taking the digit 9 as an example, under the same storage cost, the synthetic data obtained based on the method in this chapter, as shown in examples b and c, exhibit both solid and outline features of the digit 9, while the data 9 from the benchmark method only has the solid feature as shown in example a. This indicates that the synthetic dataset obtained through our method has better feature diversity compared to the benchmark method, which is a key reason for its superior meta-test accuracy.

\subsection{Hyperparameter experiments with low-rank value r}

\(r \) represents the dimension of the low-dimensional space and is a crucial hyperparameter in the experiments. The analysis is conducted on the CIFAR-10 dataset with a compression rate of 0.02\%. The variation in \( r \) also leads to changes in the number of "low-dimensional mappers" and the number of "basis vectors." Here, the value of \( r \) is selected such that the total storage cost of all components closely matches the size of the space specified by the given compression rate. The specific values are detailed in Table 2.

The visualization results of the synthesized images are presented in Fig 5. As illustrated in Fig 5, the higher the low-rank value, the closer the details of the synthesized images resemble those obtained through the baseline method. When \( r = 16 \), the synthesized images can effectively remove background noise textures while preserving the class information with minimal loss. However, reducing the low-rank value can significantly decrease the acquisition cost of the synthesized images. 

\subsection{Soft labels and progressive optimization strategy ablation experiments}
In this section, we evaluate the effectiveness of soft labels, progressive optimization, and low-rank space sampling techniques on the CIFAR-10 dataset, with the results presented in Table 3. The findings from the table indicate that the soft labeling strategy can effectively enhance test accuracy, attributed to the inter-class relationship information contained within the soft labels, which benefits the generalization performance of the network. The progressive optimization strategy performs poorly at extremely low compression rates because there are few valid trajectories to match under such conditions, often rendering the strategy unnecessary. However, as the compression rate increases and the number of learnable trajectories grows, the progressive optimization strategy not only aids the algorithm in better fitting expert trajectories but also effectively prevents potential gradient explosion issues that may arise during the initial stages of training.



%
%

\section{Conclusion}

Our paper introduces a dataset distillation algorithm leveraging low-rank space sampling. This method approximates high-dimensional data into a low-dimensional manifold space, replacing the expensive acquisition of synthetic data with sampled low-dimensional basis vectors. By significantly reducing per-image acquisition costs and enhancing dataset diversity, our approach effectively filters noise from gradient information. Minimizing storage costs per synthesized image enables the creation of extensive and diverse synthetic datasets under equivalent storage conditions. Our method offers a cost-effective solution for generating synthetic datasets that maintain data fidelity while enhancing diversity. Experimental results demonstrate the efficacy of our approach in advancing dataset distillation techniques.

\bibliographystyle{ieeenat_fullname}
\bibliography{main}
\end{document}